\documentclass[10pt,twocolumn,letterpaper]{article}

\usepackage[pagenumbers]{cvpr} 

\newtoggle{public}
\toggletrue{public}

\usepackage{multirow}
\usepackage{colortbl}
\usepackage{elasticrow}

\newbool{showText}
\boolfalse{showText}
\newcommand{\toggleText}[1]{\ifbool{showText}{{\color{gray}#1}}{}}

\makeatletter
\DeclareRobustCommand\onedot{\futurelet\@let@token\@onedot}
\def\@onedot{\ifx\@let@token.\else.\null\fi\xspace}

\def\eg{\emph{e.g}\onedot} 
\def\ie{\emph{i.e}\onedot} 
 
\def\etc{\emph{etc}\onedot} 
\def\wrt{w.r.t\onedot} 
 
\makeatother

\newcommand\blfootnote[1]{
  \begingroup
  \renewcommand\thefootnote{}\footnote{#1}
  \addtocounter{footnote}{-1}
  \endgroup
}

\definecolor{cvprblue}{rgb}{0.21,0.49,0.74}
\usepackage[pagebackref,breaklinks,colorlinks,citecolor=cvprblue]{hyperref}

\def\paperID{****} 
\def\confName{CVPR}
\def\confYear{****}

\title{Segment and Caption Anything}

\newcommand{\authorskip}{\hspace{12mm}}
\author{
 Xiaoke Huang$^{1\dagger}$ \authorskip Jianfeng Wang$^{2}$ \authorskip  Yansong Tang$^{1\ast}$ \authorskip
 Zheng Zhang$^{2}$ \\  Han Hu$^{2}$ \authorskip Jiwen Lu$^{3}$ \authorskip Lijuan Wang$^{2}$ \authorskip Zicheng Liu$^{4}$\\[3mm]
 $^1$Shenzhen International Graduate School, Tsinghua University ~~~~~~~~~ $^2$Microsoft\\
 $^3$Department of Automation, Tsinghua University ~~~~~~ $^4$Advanced Micro Devices\\
 {\tt\small \{hxk21@mails.,tang.yansong@sz.,lujiwen@\}tsinghua.edu.cn}\\ 
 {\tt\small\{jianfw,zhez,lijuanw\}@microsoft.com} ~
 {\tt\small ancientmooner@gmail.com} ~
 {\tt\small zicheliu@amd.com}
}

\begin{document}
\maketitle

\begin{abstract}
\iftoggle{public}{
\blfootnote{
$^\dagger$Work was done when the author interned at Microsoft.
}
\blfootnote{
$^\ast$Corresponding.
}
}

We propose a method to efficiently equip the Segment Anything Model (SAM) with the ability to generate regional captions.
SAM presents strong generalizability to segment anything while is short for semantic understanding.
By introducing a lightweight query-based feature mixer, 
we align the region-specific features with the embedding space of language models for later caption generation.
As the number of trainable parameters is small (typically in the order of tens of millions), it costs less computation, less memory usage, and less communication bandwidth, resulting in both fast and scalable training.
To address the scarcity problem of regional caption data, we propose to first pre-train our model on objection detection and segmentation tasks. 
We call this step weak supervision pretraining since the pretraining data only contains category names instead of full-sentence descriptions. 
The weak supervision pretraining allows us to leverage many publicly available object detection and segmentation datasets. 
We conduct extensive experiments to demonstrate the superiority of our method and validate each design choice.
This work serves as a stepping stone towards scaling up regional captioning data and sheds light on exploring efficient ways to augment SAM with regional semantics.
\iftoggle{public}{
The project page, along with the associated code, can be accessed via the following \href{https://xk-huang.github.io/segment-caption-anything/}{link}.
}

\end{abstract}
\section{Introduction}

\begin{figure}[htbp]
    \centering
    \includegraphics[width = 0.95 \linewidth]{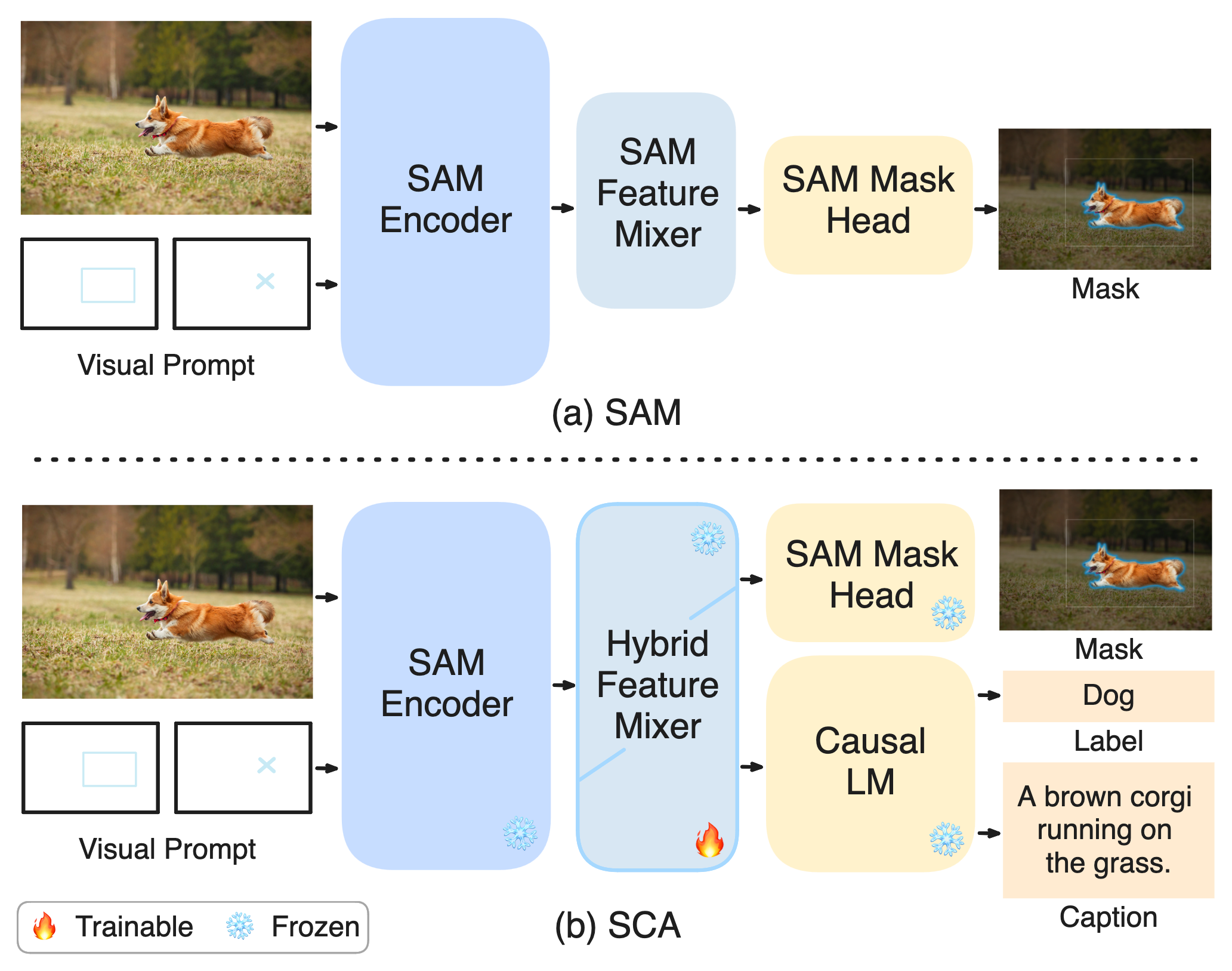}
    \caption{
    SCA (\textit{b}) is a lightweight augmentation of SAM (\textit{a}) with the ability to generate regional captions. On top of SAM architecture, we add a pre-trained language model which is frozen, and a lightweight hybrid feature mixture. Despite the small number of trainable parameters, the region-specific features are learned to align with the embedding space of the language model for regional caption generation.
    }
    \label{sec2:figs:task_motivation}
\end{figure}

Teaching machines to understand the visual world with natural languages has been a long-standing problem in computer vision~\cite{hossain2019comprehensive_image_cap_survey_1,sharma2020image_image_cap_survey_2,stefanini2022show_image_cap_survey_3}.
Image captioning is one of the topics that require the machine to perceive and describe images in human languages~\cite{karpathyDeepVisualSemanticAlignments2015b_visual_align,kulkarni2013babytalk_baby_talk}.
With the wave of deep learning~\cite{lecun2015deep_dl_1,goodfellow2016deep_dl_2}, enormous efforts~\cite{liBLIPBootstrappingLanguageImage2022_blip,liBLIP2BootstrappingLanguageImage2023_blip2,wangGITGenerativeImagetotext2022a,yu2022coca_coca} have been devoted to pushing its frontier in terms of model architectures, training data, training techniques, \etc.
However, much less work has been devoted to the regional captioning~\cite{johnsonDenseCapFullyConvolutional2015a,wuGRiTGenerativeRegiontotext2022a,longCapDetUnifyingDense2023a_CapDet,zhangGPT4RoIInstructionTuning2023a},
in which models describe the regions instead of the entire image.

Building an intelligent system that follows human intent is an emerging research topic,
as evidenced by the rapid progress of large foundation models~\cite{brownLanguageModelsAre2020_gpt3,touvronLLaMAOpenEfficient2023b_llama,radfordLearningTransferableVisual2021b_clip,jiaScalingVisualVisionLanguage2021a_align,yuanFlorenceNewFoundation2021a}.
Major breakthroughs have been made in language modeling~\cite{brownLanguageModelsAre2020_gpt3,pengInstructionTuningGPT42023_vicuna,touvronLlamaOpenFoundation2023_llama_2,alpaca}, 
where the foundation language models are fine-tuned to follow the instructions of users with both instruction supervision~\cite{pengInstructionTuningGPT42023_vicuna,alpaca} and human feedback~\cite{ouyangTrainingLanguageModels2022_Instructgpt,touvronLlamaOpenFoundation2023_llama_2}.
The idea is further developed in multi-modal language model ~\cite{liuVisualInstructionTuning2023c_llava,zhuMiniGPT4EnhancingVisionLanguage2023b}, text-to-image generation~\cite{rombach2022high_stable_diffusion,saharia2022photorealistic_imagen,yu2022scaling_parti}, and interactive segmentation~\cite{kirillovSegmentAnything2023b_SAM}.
\textit{Segment Anything Model} (SAM)~\cite{kirillovSegmentAnything2023b_SAM} is an interactive segmentation system, that successfully scales the mask data to a billion.
Such data scale enables stronger generalizability in segmentation given the visual prompts.
However, the data contain \textit{no semantic labels} thus the model is incapable of semantic understanding.

We propose a method to efficiently equip SAM with the ability to generate regional captions.
We marry SAM with causal language models~\cite{radfordLanguageModelsAre_gpt2,brownLanguageModelsAre2020_gpt3,touvronLLaMAOpenEfficient2023b_llama} by introducing a lightweight  \textit{hybrid feature mixture} which stacks a text feature mixture on top of the SAM feature mixture.
The hybrid feature mixture extracts regional features for downstream caption predictions via self- and cross-attention~\cite{vaswaniAttentionAllYou2023_transformers}.
We \textit{solely} optimize the text feature mixer and leave the other network modules (\ie SAM's encoder, SAM feature mixer, the language model) untouched.
During training, the region-specific features are aligned with the embedding space of language models for later caption generation.
As the number of trainable parameters is small (typically in the order of tens of millions), it costs less computation, less memory usage, and less communication bandwidth, resulting in both \textit{fast and scaleable} training.
\cref{sec2:figs:task_motivation} provides a system overview.

However, there is limited data available for training regional captioning models~\cite{krishnaVisualGenomeConnecting2016a_VG,yuModelingContextReferring2016a_refcoco}.
For example, One commonly used dataset, Visual Genome (VG)~\cite{krishnaVisualGenomeConnecting2016a_VG} contains up to 100K images.
In contrast, SAM~\cite{kirillovSegmentAnything2023b_SAM} used a dataset that contains more than 11M images and 1B masks.
Inspired by the effective deployment of weak supervision~\cite{liGroundedLanguageImagePretraining2022a_glipv1,zhang2023simple_openseed,zou2023generalized_xdecoder},
we introduce a weak supervision pretraining step to leverage the publicly available object detection and segmentation datasets.
Specifically, we pre-train the text feature mixer on Objects365~\cite{shaoObjects365LargeScaleHighQuality2019} detection data and COCO-Panoptic~\cite{linMicrosoftCOCOCommon2015a_mscoco} segmentation data, which consist of 1.8M images.
Finally, the text feature mixer is finetuned on the VG regional captioning data.

We have conducted extensive experiments to demonstrate the effectiveness of our method and validate each design choice.
Our method achieves state-of-the-art performance on the VG~\cite{krishnaVisualGenomeConnecting2016a_VG} benchmark with 149.8 CIDEr-D, 17.5 METEOR, and 31.4 SPICE.
We believe this work serves as a stepping stone towards scaling up regional captioning data~\cite{kirillovSegmentAnything2023b_SAM,betkerImprovingImageGeneration_dalle3,mindererScalingOpenVocabularyObject2023a_owlvitv2} and sheds light on exploring efficient approaches to augment a segmentation model like SAM with regional semantics.

\section{Related Works}

\noindent \textbf{Object detections, segmentations, and interactive segmentations}. The field of object detection has evolved from CNN-based methods~\cite{girshickRichFeatureHierarchies2014_RCNN,renFasterRCNNRealTime2016a,redmonYouOnlyLook2016_yolo,liuSSDSingleShot2016_SSD,liuSSDSingleShot2016_SSD,tianFCOSFullyConvolutional2019a,duanCenterNetKeypointTriplets2019,liu2022global, liu2022learning} to transformer-based models~\cite{vaswaniAttentionAllYou2023_transformers,carionEndtoEndObjectDetection2020b_detr,liuDABDETRDynamicAnchor2022a,zhuDeformableDETRDeformable2021,zhangDINODETRImproved2022a_dino_detr,liDNDETRAccelerateDETR2022,han2023global, luo2024soc}. The transformer architecture has shown versatility across modalities, facilitating tasks like open-world detection~\cite{radfordLearningTransferableVisual2021b_clip,jiaScalingVisualVisionLanguage2021a_align,yuanFlorenceNewFoundation2021a,zhaiSigmoidLossLanguage2023_siglip}. Similar architectural trends are observed in segmentation tasks~\cite{heMaskRCNN2018a,chengPerPixelClassificationNot2021a_maskformer,chengMaskedattentionMaskTransformer2022a_mask2former}. Recent works have also integrated vision-language pre-training for open-world segmentation~\cite{xuSideAdapterNetwork2023a_SAN,dongMaskCLIPMaskedSelfDistillation2023,liangOpenVocabularySemanticSegmentation2023_ovseg,ghiasiScalingOpenVocabularyImage2022_openseg,liLanguagedrivenSemanticSegmentation2022_lseg,yiSimpleFrameworkTextSupervised2023_simseg,liu2023universal,liu2023open}. Interactive segmentation~\cite{gradyRandomWalksImage2006,rotherGrabCutInteractiveForeground2004,rakellyFewshotSegmentationPropagation2018_DIOS} is a sub-task with unique challenges that can be tackled by transformer-based models like SAM~\cite{kirillovSegmentAnything2023b_SAM}. This paper extends SAM to region-level understanding using additional tokens and transformer layers.

\noindent \textbf{Image captioning and dense captioning}. Image captioning involves generating textual descriptions for images by combining vision and language models~\cite{dosovitskiyImageWorth16x162021a_vit,heMaskedAutoencodersAre2021a_mae,sunEVACLIPImprovedTraining2023_eva_clip,radfordLearningTransferableVisual2021b_clip,jiaScalingVisualVisionLanguage2021a_align,dingDaViTDualAttention2022,yuanFlorenceNewFoundation2021a}. Early methods employed CNN and LSTM~\cite{karpathyDeepVisualSemanticAlignments2015b_visual_align}, while recent works leverage transformers~\cite{alayracFlamingoVisualLanguage2022a,chowdheryPaLMScalingLanguage2022b_palm,liBLIP2BootstrappingLanguageImage2023_blip2,daiInstructBLIPGeneralpurposeVisionLanguage2023,wangGITGenerativeImagetotext2022a} and large language models~\cite{brownLanguageModelsAre2020_gpt3,touvronLLaMAOpenEfficient2023b_llama,touvronLlamaOpenFoundation2023_llama_2,pengInstructionTuningGPT42023_vicuna}. These models can follow user instructions, demonstrating abilities like visual reasoning and question answering. Dense captioning~\cite{krishnaVisualGenomeConnecting2016a_VG,johnsonDenseCapFullyConvolutional2015a,liLearningObjectContext2019_COCG_COCD_ImgG,yinContextAttributeGrounded2019a_CAG_Net,yangDenseCaptioningJoint2017_JIVC,shaoRegionObjectRelationAwareDense2022_TDC,longCapDetUnifyingDense2023a_CapDet,wuGRiTGenerativeRegiontotext2022a} extends image captioning to region-level, combining detection with generation. Despite its simultaneous development with image captioning, its evaluation metrics improvement has been slow due to the compounded difficulty of localization and generation. This work assumes localization proposals as given inputs and focuses on region captioning.

\begin{figure*}[t]
    \centering
    \includegraphics[width = 0.9\linewidth]{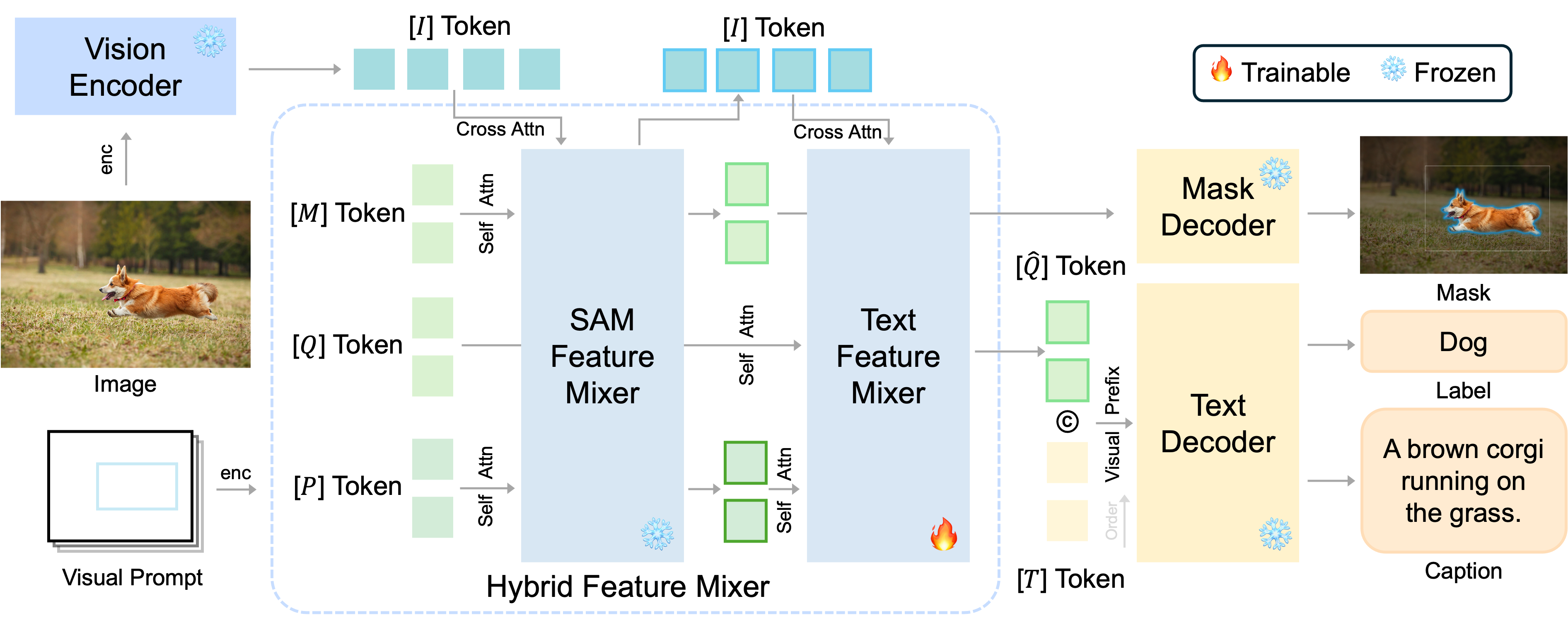}
    \caption{
    \textbf{The model architecture}.
The model consists of three parts including an image encoder, a feature mixer, and decoder heads for masks or text.
The key ingredient of the model is the \textit{text feature mixer}, which is a lightweight bidirectional transformer~\cite{vaswaniAttentionAllYou2023_transformers}.
We stack it over the one from SAM and reuse its tokens.
By solely optimizing the additional mixer, we align the region-specific features with the embedding space of language models.
The training is both fast and scalable thanks to the limited amount of optimizable parameters.
    }
    \label{sec3:model:model}
\end{figure*}

\noindent \textbf{Scaling region understanding systems}. Tremendous progress has been made in natural language processing and vision domains by training large models on massive datasets, with scaling laws illustrating the relationship between computational budgets, data size, and performance~\cite{hoffmannTrainingComputeOptimalLarge2022b_Chinchilla,radfordLearningTransferableVisual2021b_clip,jiaScalingVisualVisionLanguage2021a_align,yuanFlorenceNewFoundation2021a,alayracFlamingoVisualLanguage2022a,chowdheryPaLMScalingLanguage2022b_palm,anilPaLMTechnicalReport2023_palm2,openaiGPT4TechnicalReport2023}. This trend is also observed in region-level understanding systems, where weak-supervision methods like self-training and proxy training losses are used to scale up data~\cite{zhongRegionCLIPRegionbasedLanguageImage2021,zhouDetectingTwentythousandClasses2022_detic,arandjelovicThreeWaysImprove2023threeways,liGroundedLanguageImagePretraining2022a_glipv1,zhangGLIPv2UnifyingLocalization2022a,yaoDetCLIPDictionaryEnrichedVisualConcept2022,yaoDetCLIPv2ScalableOpenVocabulary2023,he2023camouflaged,he2024strategic,he2024weakly}. \cite{mindererScalingOpenVocabularyObject2023a_owlvitv2} and \cite{kirillovSegmentAnything2023b_SAM} show the importance of scaling in vision tasks by reaching the scale of billions of samples. However, region-level understanding is still underexplored due to the limited data scale. The current dataset, Visual Genome~\cite{krishnaVisualGenomeConnecting2016a_VG}, is small, leading to poor alignment and generalizability. This work aims to explore the scaling property in generative region-level understanding using weak supervision from detection ~\cite{linMicrosoftCOCOCommon2015a_mscoco,shaoObjects365LargeScaleHighQuality2019,kuznetsovaOpenImagesDataset2020_openimage,wangV3DetVastVocabulary2023a} and leaves image captioning supervision~\cite{,ordonezIm2TextDescribingImages2011_sbu_datast,sharmaConceptualCaptionsCleaned2018_cc3m,changpinyoConceptual12MPushing2021_cc12m} and self-training~\cite{mindererScalingOpenVocabularyObject2023a_owlvitv2,liBLIPBootstrappingLanguageImage2022_blip} for future exploration.

\noindent \textbf{Concurrent works}. Recent progress in Large Language Model (LLM) and interactive segmentation has spurred several concurrent works in region-level understanding. Without training, Caption Anything~\cite{wangCaptionAnythingInteractive2023a} utilizes SAM and image captioning models to predict text descriptions based on the cropped regions, with style adjustments by ChatGPT~\cite{ouyangTrainingLanguageModels2022_Instructgpt}. Other works train with existing data; GPT4ROI~\cite{zhangGPT4RoIInstructionTuning2023a} extends Visual LLM~\cite{liuVisualInstructionTuning2023c_llava} to process region prompts, while Region-BLIP~\cite{zhouRegionBLIPUnifiedMultimodal2023a} employs BLIP2's feature mixer~\cite{liBLIP2BootstrappingLanguageImage2023_blip2} and trains on multiple tasks. Works like Kosmos-2~\cite{pengKosmos2GroundingMultimodal2023a} and All-Seeing~\cite{wangAllSeeingProjectPanoptic2023} utilize similar architectures but different dataset construction paths, demonstrating strong performance on various region-level tasks. Despite the rapid evolution of this field, this work aims to extend SAM for region-level captioning with weak supervision.

\section{Method}

There are three components in the model, a ViT-based encoder, a transformer query-based feature mixer, and decoder heads for different outputs of interest, \eg text decoder.
Our model design is inspired by~\cite{kirillovSegmentAnything2023b_SAM}, which is a category-agnostic promptable segmentation model that takes in user inputs like points, boxes, or masks and outputs multiple binary masks.
Apart from a ViT-based encoder~\cite{dosovitskiyImageWorth16x162021a_vit,liExploringPlainVision2022c_vitdet} and a small mask decoder~\cite{chengMaskedattentionMaskTransformer2022a_mask2former,chengPerPixelClassificationNot2021a_maskformer},
it involves a lightweight query-based feature mixer~\cite{carionEndtoEndObjectDetection2020b_detr} to mix both global image features extracted by the image encoder and the user prompts.
The module is efficient as it only consists of 2M of parameters.
\cref{sec3:model:model} illustrate the model architecture.

The data used to train the SAM model is category agnostic and after initial human labeling, the data are scaled to 10M images and 1B boxes with several rounds of self-training.
Although initially, the labeled masks involve no textual labels, they contain the semantics \textbf{implicitly} as the annotators are asked to draw masks to whatever things or stuff they recognize.
Thus we hypothesize that the features from the image encoder of SAM contain rich semantic features beyond the lower-level segmentation tasks it is trained on.
Based on that assumption, we build our model over the pre-trained SAM models and stack an additional feature mixer along with a text decoder to predict texts.
We follow the mixer design of SAM~\cite{kirillovSegmentAnything2023b_SAM} except for increasing the number of layers in the mixer.

\noindent \textbf{Image encoder}. Following SAM, a ViT style image encoder~\cite{dosovitskiyImageWorth16x162021a_vit} that is designed for detection~\cite{liExploringPlainVision2022c_vitdet} is adopted in our model.
Specifically, it is comprised of a plain ViT with primary local window attention and several interleaved global attention,
which produces isotropic feature maps with the same feature dimension.

Given image $\mathcal{I}$, we have the encoder $E_I$ that extract the global image feature $I$: $E_I(\mathcal{I}) = {I}$.
The image features are down-sampled for computation efficiency, as the following feature mixer should be as lightweight as possible.
Following~\cite{liExploringPlainVision2022c_vitdet,kirillovSegmentAnything2023b_SAM}, the final spatial shape and feature dimension are $64\times64$ and $256$, respectively (\cref{sec3:model:model}).

Not that we only utilize the single level of visual features from the last layer as in~\cite{kirillovSegmentAnything2023b_SAM}, compared with~\cite{renFasterRCNNRealTime2016a,heMaskRCNN2018a,duanCenterNetKeypointTriplets2019,redmonYouOnlyLook2016_yolo,linFocalLossDense2020_retinanet} that produce multi-scale features.
However, the single-level feature contains sufficient information for later caption generation, regardless of the scales of the regions.

\noindent \textbf{Regional feature mixer}. 
After the global image features are extracted, we need to further extract the region features denoted by the user-input visual prompts.
There are two prominent approaches to devising a region feature mixer to attain the region-of-interest (ROI) features.
The first one leverages the ROI-align operator~\cite{heMaskRCNN2018a}, which pools the region features from the global ones with the corresponding box coordinates.
The second one utilizes the attention mechanism~\cite{vaswaniAttentionAllYou2023_transformers} by incorporating query tokens that fuse the feature of interest across each attention block.
We choose the latter giving the following considerations:
1) Versatile encoding of visual prompts. 
The type of visual prompts could be either point, stroke, box, mask, or a combination of any of them.
The ROI-align operator only takes box prompts,
while in the query-based token mixer, we can encode the different formats of prompts with specific prompt encoders,
whose outputs are tokens that are compatible with the latter attention blocks.
2) Progressive feature interaction and fusion.
The main body of the query-based feature mixer is attention blocks as in~\cite{chengMaskedattentionMaskTransformer2022a_mask2former,carionEndtoEndObjectDetection2020b_detr},
whose inputs are the encoded prompt tokens, global image tokens, and task-oriented query tokens.
After several blocks of self-attentions and cross-attentions, we can fetch the region features at the exact position of the query tokens.
Unlike the process of the ROI-align operator, which only pools the global image features,
the query-based one can leverage the powerful attention mechanism to extract region-specific features that facilitate the downstream tasks,
\eg segmentation, captioning, \etc.

Given the global image tokens $I$, and user-provided prompts $\mathcal{P}_{\{b,p,m\}}$ in forms of box $b$, point $p$, or mask $m$,
we first encode the given prompts with the corresponding prompt encoders $E_p$ by $E_p( \mathcal{P}_{\{b,p,m\}} ) = P_{\{b,p,m\}}$,
where $P_{\{b,p,m\}}$ is encoded prompt tokens.
Next, we concatenate the encoded prompt tokens and both the textual and mask query tokens $Q$ and $M$,
and feed them with the global image tokens $I$ into the query-based feature mixer $E_R$ with $N$ blocks:
\begin{equation}
    E_R^j( P^{j-1}, Q^{j-1}, M^{j-1}; I^{j-1}) = \{\hat P^j, \hat Q^j, \hat M^j; \hat I^j\},
\end{equation}
where $j=\{1,2,\ldots,N\}$ is the block indicator, $\{\hat P^j, \hat Q^j, \hat M^j; \hat I^j\}$ are the fused tokens after the $j$-th block, $\{\hat P^0, \hat Q^0, \hat M^0; \hat I^0\}$ is the initial input tokens.
We denote $\{\hat P^N, \hat Q^N, \hat M^N; \hat I^N\} = \{\hat P, \hat Q, \hat M; \hat I\}$ as the final outputs.
The encoded query tokens $\hat Q$ and $\hat M$ are deemed as the ROI tokens for captioning and segmentation, respectively, which are delivered to the following output heads (\ie,  the text generation head and the mask prediction).

The query-based feature mixer $E_R$ is a bi-directional transformer with stack of blocks as in~\cite{kirillovSegmentAnything2023b_SAM,carionEndtoEndObjectDetection2020b_detr,vaswaniAttentionAllYou2023_transformers,chengMaskedattentionMaskTransformer2022a_mask2former}.
Each block consists of one self-attention layer to fuse the sparse tokens (\ie, the concatenated tokens of the prompt ones $P$ and the query ones $Q$), and a cross-attention layer to instill the global image tokens $I$.
During the encoding process across each block, the query tokens $Q$ can gradually gather the task-specific information grounded by the prompts ones $P$, 
inside the global image tokens $I$.

\noindent \textbf{Query tokens}.
\cite{kirillovSegmentAnything2023b_SAM} takes query-based feature mixer as its core component but only predicts the masks without high-level semantic outputs like labels.
We notice that~\cite{kirillovSegmentAnything2023b_SAM} can actually predict masks with good semantics even if it is trained by a category-agnostic approach.
It may be attributed to the initial training data of SAM, which are labeled under the instruction where
the annotators are asked to draw the masks over whatever things of stuff they recognized without any semantic labels.
Thus we leverage the query tokens from~\cite{kirillovSegmentAnything2023b_SAM} by stacking an additional feature mixer $E_R^\text{Cap}$ above that of~\cite{kirillovSegmentAnything2023b_SAM}.
Specifically, \cite{kirillovSegmentAnything2023b_SAM} possessed its own query tokens $M$ to mix the features for mask prediction.
It encoded the corresponding features with a two-layer feature mixer $E_R^\text{SAM}$.
We add a new set of query tokens $Q$ for text predictions and feed it with the prompt tokens and image tokens that are both encoded with  $E_R^\text{SAM}$ into $E_R^\text{Cap}$.

\begin{table*}[tbhp]
\centering
\caption{
Comparison with baselines.
``C'': CIDEr-D~\cite{vedantam2015cider}, ``M'': METEOR~\cite{banerjee2005meteor}, ``S'': SPICE~\cite{anderson2016spice}, ``B'': BLEU~\cite{papineni2002bleu}, ``R'': ROUGE~\cite{lin2004rouge},
``(F)'': Fuzzy.
For all metrics, the higher the better.
The best, the second best, the third best scores are marked as \colorbox[rgb]{1,0.698,0.698}{\strut red}, \colorbox[rgb]{1,0.851,0.698}{\strut orange}, \colorbox[rgb]{0.996,1,0.698}{\strut yellow}, respectively.
$\ast$: The captioners used in~\cite{wangCaptionAnythingInteractive2023a}.
$\dagger$: We pre-train the model for 100K steps, then finetune it on VG for 100K steps.
$\ddagger$: When no pertaining is applied, we train the model on VG for 200K steps.
Thus they have similar training costs.
}
\label{sec4:tab:comp}
\scalebox{0.87}{
\begin{tabular}{lcccccccccccc} 
\toprule
Method             & {C}                   & {M}                  & {S}                  & {B@1}                & {B@2}                & {B@3}                & {B@4}                & {R}                  & {Noun}               & {Verb}              & {Noun (F)}           & {Verb (F)}            \\ 
\hline
SAM+BLIP-base           & 43.8                                  & 9.6                                  & 12.6                                 & 16.8                                 & 7.8                                  & 3.9                                  & 2.1                                  & 19.8                                 & 21.4                                 & 3.0                                 & 49.6                                 & 8.2                                   \\
SAM+BLIP-large$^\ast$          & 25.3                                  & 11.0                                 & 12.7                                 & 14.1                                 & 6.5                                  & 3.2                                  & 1.6                                  & 18.5                                 & 27.3                                 & 4.3                                 & 56.2                                 & {\cellcolor[rgb]{1,0.698,0.698}}12.4  \\
SAM+GIT-base            & 65.5                                  & 10.1                                 & 17.1                                 & 23.6                                 & 11.7                                 & 7.1                                  & 4.8                                  & 21.8                                 & 22.7                                 & 1.4                                 & 49.8                                 & 3.0                                   \\
SAM+GIT-base-coco       & 67.4                                  & 11.2                                 & 17.5                                 & 24.4                                 & 12.6                                 & 7.5                                  & 4.9                                  & 23.1                                 & 25.6                                 & 2.5                                 & 52.7                                 & 5.2                                   \\
SAM+GIT-base-textcaps   & 45.6                                  & 11.6                                 & 15.0                                 & 18.4                                 & 8.9                                  & 4.7                                  & 2.7                                  & 21.8                                 & 26.1                                 & 3.5                                 & 54.2                                 & 7.4                                   \\
SAM+GIT-large$^\ast$           & 68.8                                  & 10.5                                 & 17.8                                 & 24.2                                 & 12.3                                 & 7.4                                  & 5.0                                  & 22.4                                 & 24.5                                 & 1.8                                 & 51.6                                 & 3.7                                   \\
SAM+GIT-large-coco      & 71.8                                  & 12.2                                 & 18.8                                 & 24.6                                 & 12.9                                 & 7.7                                  & 4.9                                  & 24.4                                 & 28.9                                 & 3.4                                 & 55.8                                 & 6.7                                   \\
SAM+GIT-large-textcaps  & 59.2                                  & 12.6                                 & 17.5                                 & 20.9                                 & 10.5                                 & 6.0                                  & 3.6                                  & 23.6                                 & 29.4                                 & 3.7                                 & 56.5                                 & 7.2                                   \\
SAM+BLIP2-OPT-2.7B-coco & 30.4                                  & 11.3                                 & 12.0                                 & 14.4                                 & 7.1                                  & 3.6                                  & 1.9                                  & 19.3                                 & 26.7                                 & {\cellcolor[rgb]{0.996,1,0.698}}4.7 & 55.0                                 & {\cellcolor[rgb]{1,0.851,0.698}}12.1  \\
SAM+BLIP2-OPT-2.7B$^\ast$      & 59.7                                  & 11.7                                 & 16.7                                 & 19.6                                 & 9.8                                  & 5.3                                  & 3.0                                  & 22.7                                 & 26.6                                 & 4.5                                 & 53.7                                 & 9.7                                   \\
SAM+BLIP2-OPT-6.7B-coco & 30.4                                  & 12.2                                 & 13.1                                 & 14.7                                 & 7.3                                  & 3.8                                  & 2.0                                  & 19.9                                 & 29.7                                 & {\cellcolor[rgb]{0.996,1,0.698}}4.7 & 57.8                                 & {\cellcolor[rgb]{0.996,1,0.698}}11.7  \\
SAM+BLIP2-OPT-6.7B      & 56.6                                  & 11.7                                 & 16.2                                 & 19.0                                 & 9.5                                  & 5.0                                  & 2.8                                  & 22.3                                 & 26.7                                 & 4.4                                 & 53.9                                 & 10.1                                  \\ 
\hline
GRiT                    & 142.2                                 & 17.2                                 & 30.5                                 & 36.0                                 & 22.1                                 & 15.2                                 & 11.2                                 & 34.5                                 & 39.5                                 & 4.3                                 & 63.3                                 & 7.2                                   \\ 
\hline
SCA (GPT2-large, VG)$^\ddagger$          & {\cellcolor[rgb]{0.996,1,0.698}}148.8 & {\cellcolor[rgb]{0.996,1,0.698}}17.4 & {\cellcolor[rgb]{0.996,1,0.698}}31.2 & {\cellcolor[rgb]{1,0.851,0.698}}38.0 & {\cellcolor[rgb]{1,0.851,0.698}}23.9 & {\cellcolor[rgb]{0.996,1,0.698}}16.6 & {\cellcolor[rgb]{0.996,1,0.698}}12.1 & {\cellcolor[rgb]{1,0.851,0.698}}35.5 & {\cellcolor[rgb]{1,0.851,0.698}}41.5 & {\cellcolor[rgb]{1,0.698,0.698}}4.8 & {\cellcolor[rgb]{1,0.851,0.698}}65.0 & 7.6  \\
SCA (LLAMA-3B, VG)$^\ddagger$            & {\cellcolor[rgb]{1,0.851,0.698}}149.8 & {\cellcolor[rgb]{1,0.851,0.698}}17.4 & {\cellcolor[rgb]{1,0.851,0.698}}31.3 & {\cellcolor[rgb]{0.996,1,0.698}}38.0 & {\cellcolor[rgb]{0.996,1,0.698}}23.9 & {\cellcolor[rgb]{1,0.851,0.698}}16.7 & {\cellcolor[rgb]{1,0.698,0.698}}12.2 & {\cellcolor[rgb]{0.996,1,0.698}}35.5 & {\cellcolor[rgb]{0.996,1,0.698}}41.2 & 4.5                                 & {\cellcolor[rgb]{0.996,1,0.698}}64.6 & 7.1  \\
SCA (GPT2-large, Pretrain+VG)$^\dagger$ & {\cellcolor[rgb]{1,0.698,0.698}}149.8 & {\cellcolor[rgb]{1,0.698,0.698}}17.5 & {\cellcolor[rgb]{1,0.698,0.698}}31.4 & {\cellcolor[rgb]{1,0.698,0.698}}38.2 & {\cellcolor[rgb]{1,0.698,0.698}}24.1 & {\cellcolor[rgb]{1,0.698,0.698}}16.8 & {\cellcolor[rgb]{1,0.851,0.698}}12.2 & {\cellcolor[rgb]{1,0.698,0.698}}35.7 & {\cellcolor[rgb]{1,0.698,0.698}}41.7 & {\cellcolor[rgb]{1,0.851,0.698}}4.8 & {\cellcolor[rgb]{1,0.698,0.698}}65.1 & 7.5  \\
\bottomrule
\end{tabular}
}
\end{table*}

\looseness=-1
\noindent \textbf{Regional feature decoder}. 
After obtaining the ROI feature, we can send it into a causal text decoder~\cite{radfordLanguageModelsAre_gpt2,brownLanguageModelsAre2020_gpt3,touvronLLaMAOpenEfficient2023b_llama,touvronLlamaOpenFoundation2023_llama_2} to generate region captions.
The text decoder $D_\text{Cap}$ is often a transformer decoder~\cite{vaswaniAttentionAllYou2023_transformers} that predict the text tokens $\mathcal{T}_k$ based on the previous (predicted) text tokens $\mathcal{T}_{1:k-1}$ causally:
\begin{equation}
    D_\text{Cap}(\mathcal{T}_{1:k-1}) = \mathcal{T}_k,
\end{equation}
where $k$ is the length of the text tokens.
Since we want to condition the prediction on the region features, we prefix the feature token $Q$ in front of the text tokens $\mathcal{T}$.
Inspired by prompt tuning~\cite{jiaVisualPromptTuning2022_VPT,zhouLearningPromptVisionLanguage2022a_coop,liPrefixTuningOptimizingContinuous2021_prefix_tuning}, we further prefix a set of optimizable task tokens $T$ to exploit task related context (\cref{sec3:model:model}).
The model can be optimized by minimizing cross-entropy loss $L$ defined on the next token by:
\begin{equation}
    L = \frac{1}{N_\mathcal{T} + 1} \sum_{k=1}^{N_\mathcal{T} + 1} \text{CE} (\mathcal{T}_k, p( \mathcal{T}_k | T, Q, \mathcal{T}_{0:k-1})),
\end{equation}
where $p( \mathcal{T}_k | T, Q, \mathcal{T}_{1:k-1})$ is the predicted logits for token $\mathcal{T}_k$
$N_\mathcal{T}$ is the length until the predicted tokens and $N_r$ is the length of the prefix tokens.
$\text{CE}$ is cross entropy loss with label smoothing at a strength of $0.1$.
$\mathcal{T}_{0}, \mathcal{T}_{N_\mathcal{T} + 1}$ are the begin-of-sentence (BOS) and end-of-sentence (EOS) tokens, respectively.
For the details of the mask decoder, please refer to \cite{kirillovSegmentAnything2023b_SAM}.

\section{Experiments}

\subsection{Implementation Details}

Our model is comprised of three parts: image encoder, regional feature mixer, and regional feature decoder.
The image encoder is the pre-trained ViT-base or -large from~\cite{kirillovSegmentAnything2023b_SAM}.
The mask feature mixer along with mask query tokens $M$ and the mask decoder are from the pre-trained SAM.
For the text decoder, we leverage the pre-trained language model such as GPT2-large~\cite{radfordLanguageModelsAre_gpt2} and OpenLLAMA-3B~\cite{touvronLLaMAOpenEfficient2023b_llama,computerRedPajamaDataOpenSource2023_RedPajama,gengOpenLLaMAOpenReproduction2023}.
The above modules are all \textit{fixed} during training.
As to the additional transformer region feature mixer to extract textual features, we scale the 2-layer one in~\cite{kirillovSegmentAnything2023b_SAM} to 12 layers.
The caption query tokens $Q$ have a length of 8 and the task tokens $T$ have a length of 6.
We \textit{optimize} the above modules for region captioning generation.
Note that only a small set of parameters are optimized, thus the training is not only scalable but efficient.
We list the hyper-parameters in supplementary.
We first pre-train the model for 100K steps, with Objects365~\cite{shaoObjects365LargeScaleHighQuality2019} (detection) and COCO-Panoptic~\cite{linMicrosoftCOCOCommon2015a_mscoco} (segmentation) with a sampling ratio of 10:1.
Then we fine-tune the model on Visual Genome~\cite{krishnaVisualGenomeConnecting2016a_VG} dense caption split for another 100K steps.
Meanwhile, we also directly train the models on VG for 200K steps.
For inference, we use a beam size of 3 for text generation.
Note that as only the lightweight text feature mixer is optimized, we can \textit{switch} it during inference to generate either class labels (from pertaining) or captions (from finetuning).
We list more details in the supplementary materials.

\begin{figure*}
    \centering
    \begin{elasticrow}[1.7em]
        \centering
        \elasticfigure{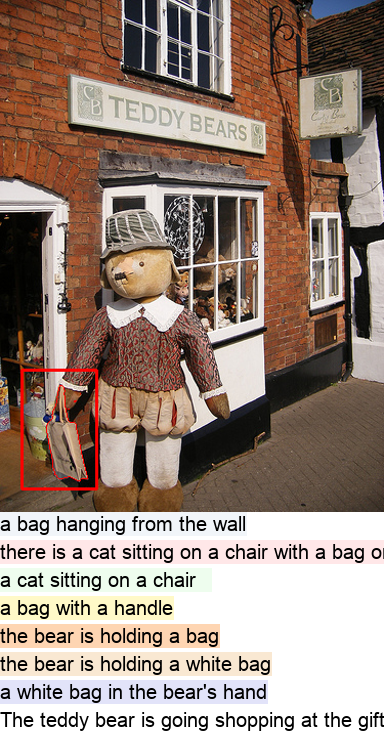}
        \elasticfigure{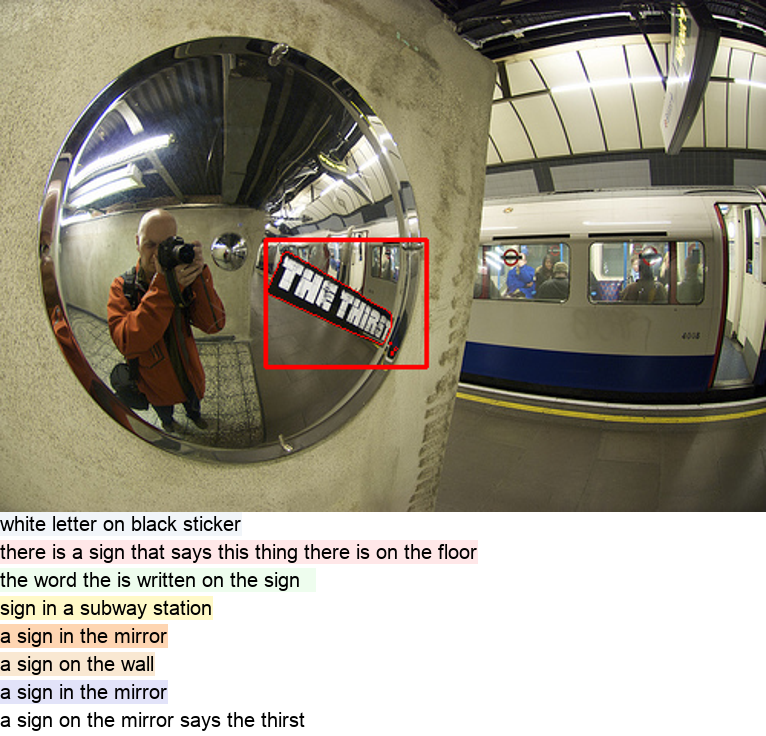}
        \elasticfigure{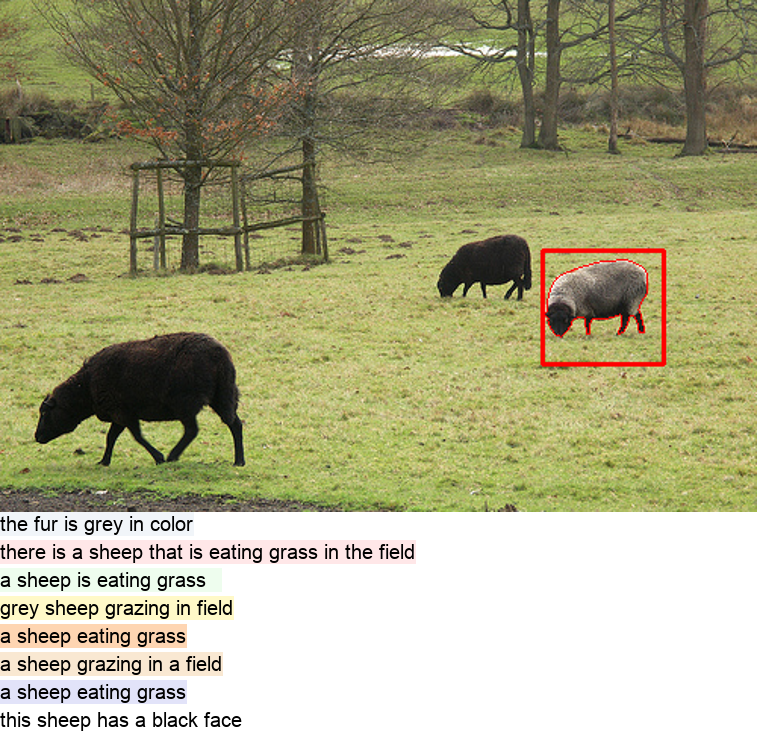}
        \elasticfigure{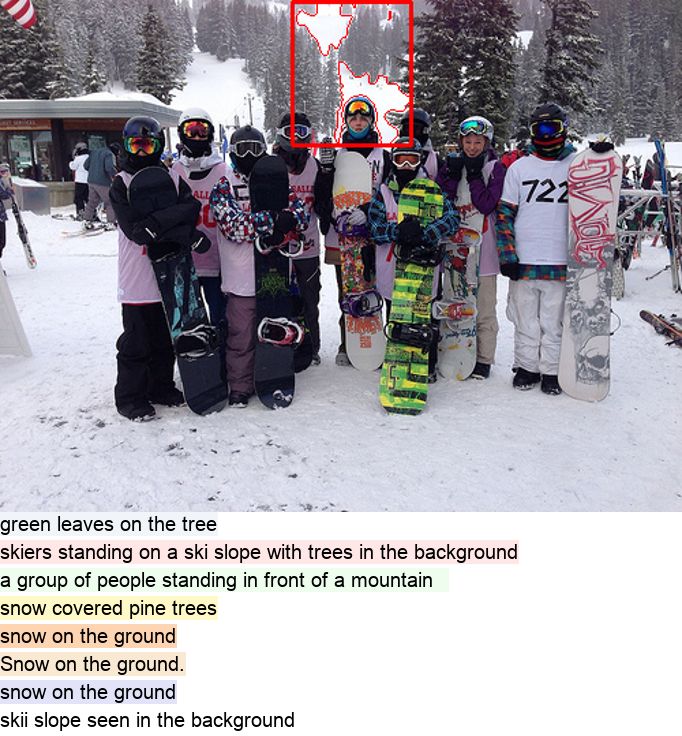}
    \end{elasticrow}

    \begin{elasticrow}[1.7em]
        \elasticfigure{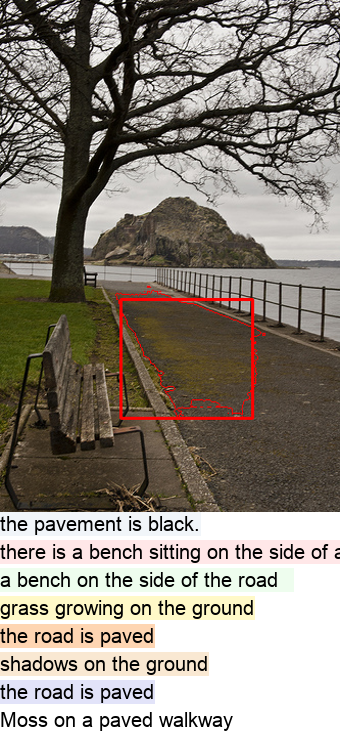}
        \elasticfigure{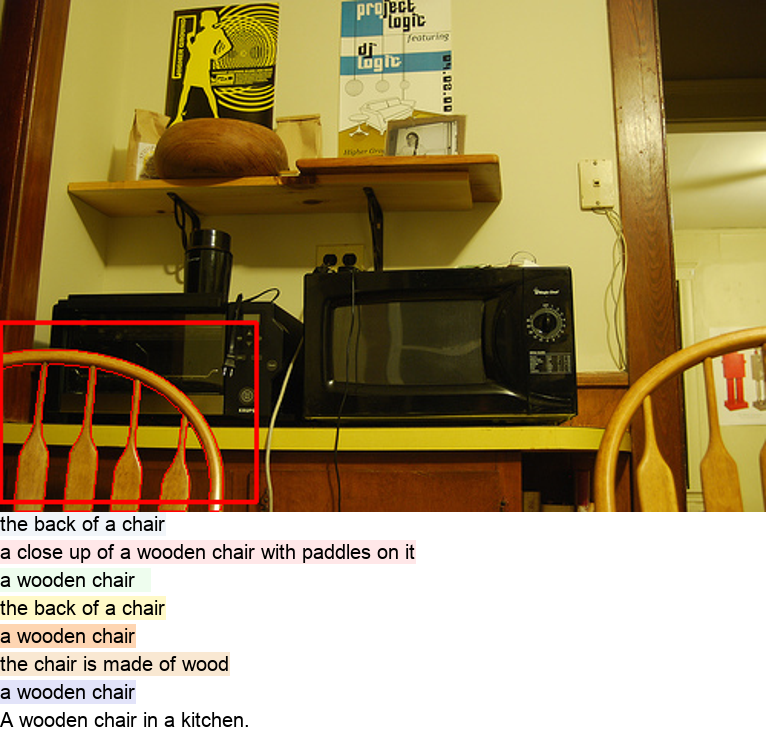}
        \elasticfigure{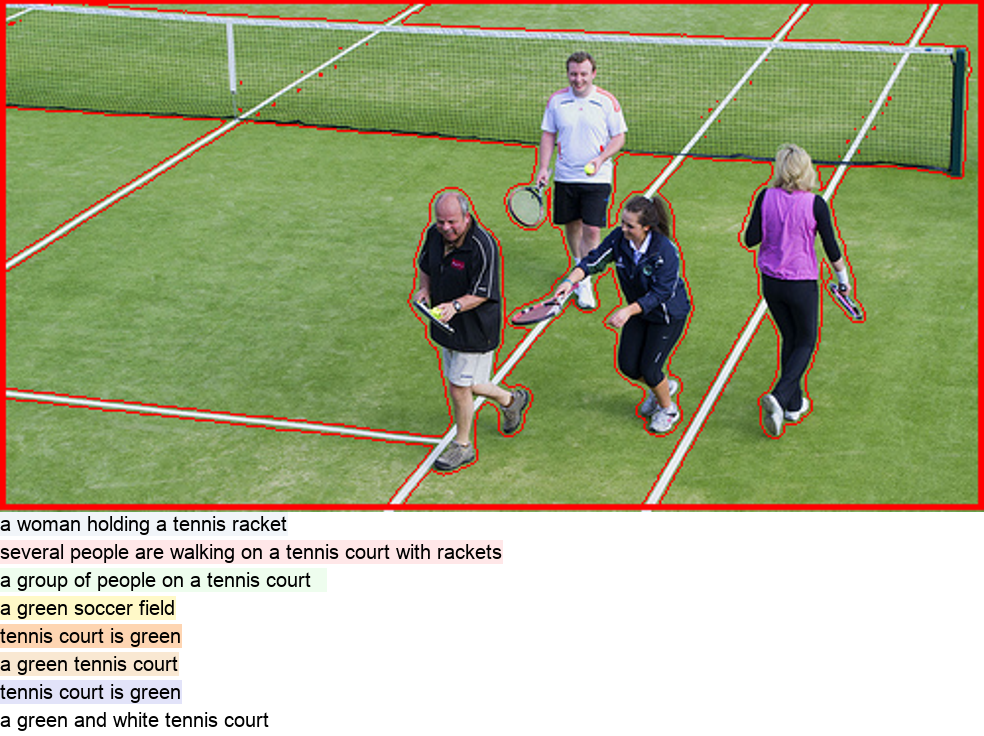}
        \elasticfigure{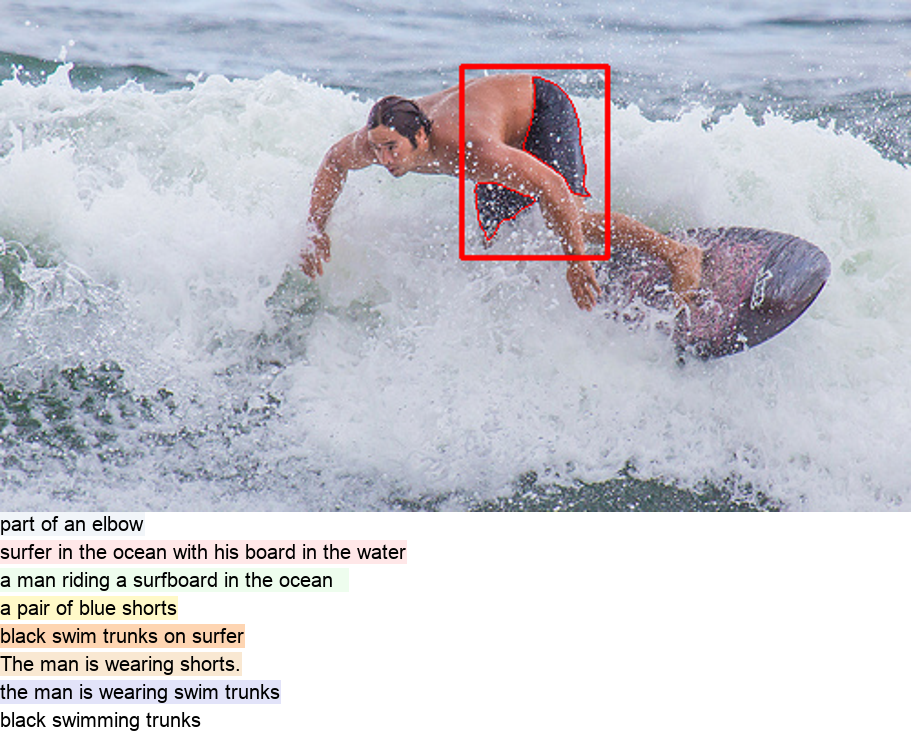}
    \end{elasticrow}

    \begin{elasticrow}[1.7em]
        \elasticfigure{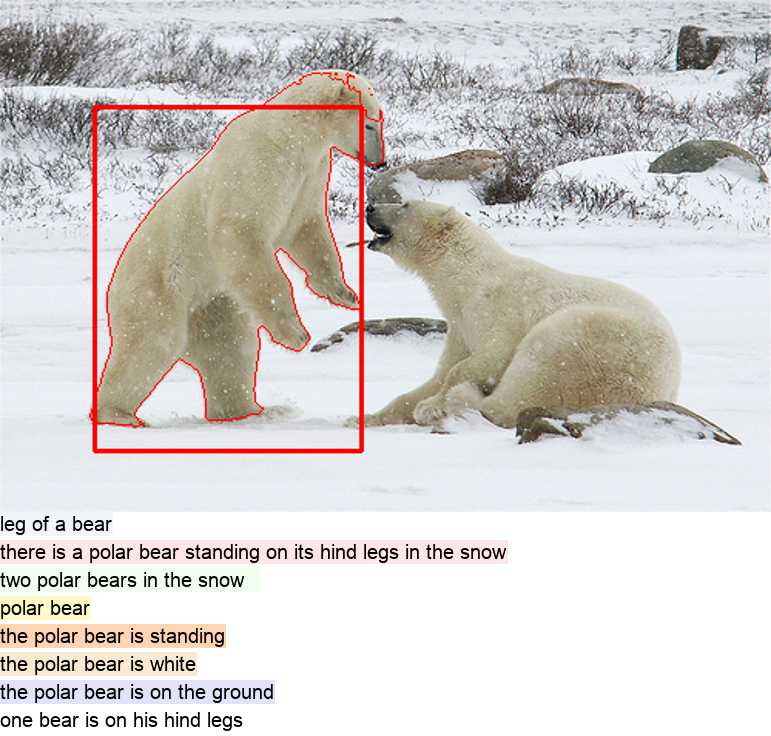}
        \elasticfigure{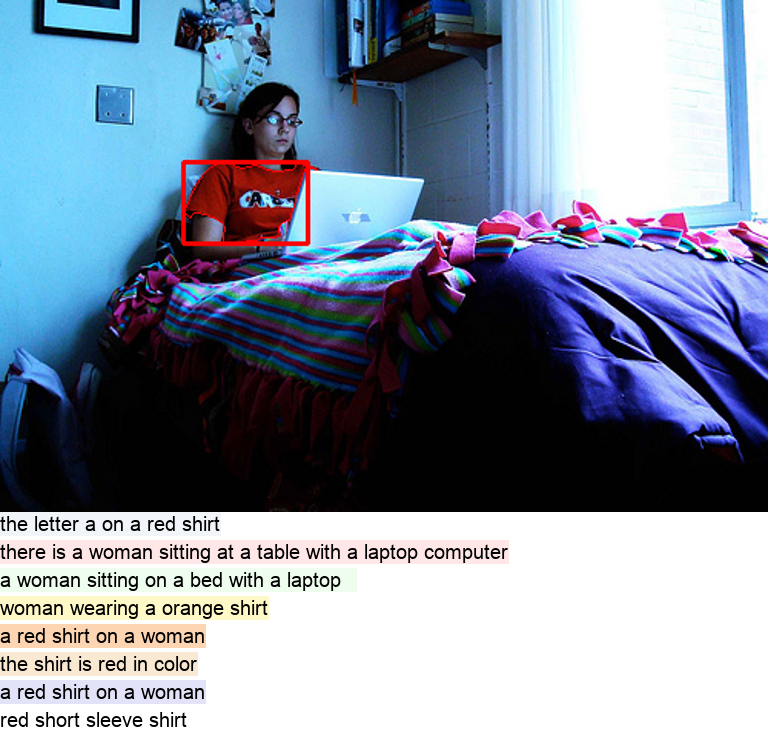}
        \elasticfigure{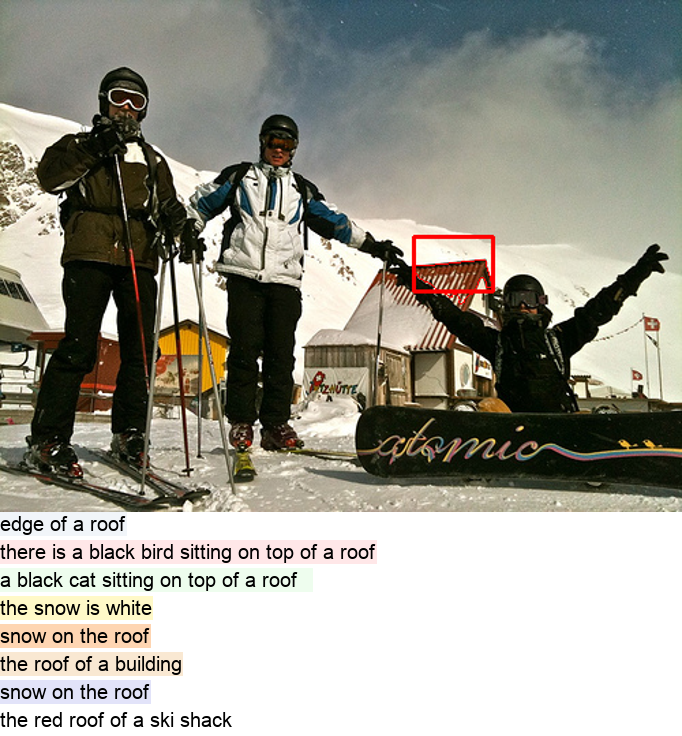}
        \elasticfigure{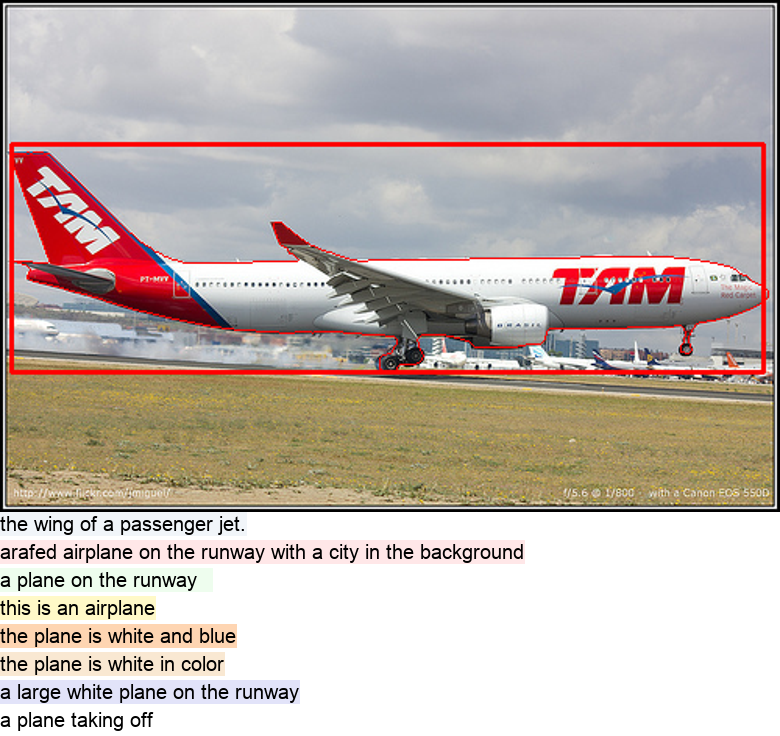}
    \end{elasticrow}

    \caption{
        The qualitative results.
        SCA simultaneously predicts masks (in red contour) and captions.
        From top-to-bottom, the captions are from: \textit{SAM+Captioner}~\{\colorbox[rgb]{0.945, 0.960, 0.976}{GIT-large}, \colorbox[rgb]{1.0, 0.905, 0.909}{BLIP-large}, \colorbox[rgb]{0.933, 0.992, 0.933}{BLIP2-OPT-2.7B}\}~\cite{wangCaptionAnythingInteractive2023a}, \colorbox[rgb]{1.0, 0.976, 0.784}{GRIT}~\cite{wuGRiTGenerativeRegiontotext2022a}, \textit{SCA}~\{\colorbox[rgb]{0.996, 0.835, 0.698}{GPT2-large+VG}, \colorbox[rgb]{0.976, 0.909, 0.823}{LLAMA-3B+VG}, \colorbox[rgb]{0.886, 0.890, 0.976}{GPT2-large+Pretrain+VG}\}, and the \textit{ground truth}.
        The bounding boxes (in red) are used to prompt the models.
        Zoom in for a better view.
    }
    \label{sec4:fig:qualitative}
\end{figure*}

\subsection{Evaluation Settings}

\noindent \textbf{Datasets}. 
We evaluate the methods on Visual Genome (VG)~\cite{krishnaVisualGenomeConnecting2016a_VG} captioning splits.
It contains about 100K images along with around 3M regions, and each region contains one textual description.
Despite the large scale of regions, there are a large number of repeated annotations due to its data curation.
We take the standard data split protocol~\cite{johnsonDenseCapFullyConvolutional2015a,yangDenseCaptioningJoint2017_JIVC,wuGRiTGenerativeRegiontotext2022a}, in which around 70K images are used for training, and other 5K images are used for evaluation.
Compared with previous works~\cite{johnsonDenseCapFullyConvolutional2015a,wuGRiTGenerativeRegiontotext2022a}, we do not preprocess the text (e.g., case conversion, remove the symbols, \etc),
as we find no performance degradation thanks to the employment of pre-trained language models.

\noindent \textbf{Metrics}.
We adopt the standard referring-based text similarity measurements~\cite{vedantam2015cider,banerjee2005meteor,anderson2016spice,papineni2002bleu,lin2004rouge} used in image captioning~\cite{wangGITGenerativeImagetotext2022a,liBLIP2BootstrappingLanguageImage2023_blip2,yu2022coca_coca},
to evaluate the generated regional captions against the ground-truth ones.
Unlike dense captioning task~\cite{johnsonDenseCapFullyConvolutional2015a,yangDenseCaptioningJoint2017_JIVC} which considers both localization and generation,
we assume localization proposals as given inputs and focus on region captioning.
Moreover, we evaluate the concepts learned by the models with phrase coverage rate.
We parse both sentences into phrases and then compute the coverage score via Intersection Over Union (IoU) for both nouns and verbs~\cite{chan2023ic}. 
The score for each pair is either exact matching or fuzzy matching, \ie the cosine similarity between the phrase embeddings.
Finally, we average the scores across all samples.

\subsection{Comparison with Other Methods}

We compare our methods with two kinds of baselines on the test split of VG.
The first baseline is \textit{training-free}, which is a SAM followed by an image captioner~\cite{liBLIPBootstrappingLanguageImage2022_blip,liBLIP2BootstrappingLanguageImage2023_blip2,wangGITGenerativeImagetotext2022a}.
It is the major algorithm in Caption Anything~\cite{wangCaptionAnythingInteractive2023a}.
We evaluate various open-sourced captioners;
The second baseline is the GRiT model~\cite{wuGRiTGenerativeRegiontotext2022a}, which is trained on the train split of VG like ours.
However, it contains a region generator to automatically generate region proposals, while ours requires those from users.
We directly test its captioning ability by providing ground truth boxes.

\cref{sec4:tab:comp} demonstrates the superior results of our models.
The image captioner baselines yield the least performance.
We speculate that the image patches generated by SAM lose the context information, 
and they differ from the training distribution of the captions \wrt both resolution and semantics.
Thus it could generate captions that are either misinformative or unspecific.
The second baseline, GRiT, gives the most competitive results, but it possesses major drawbacks in comparison with ours.
1) The full model of GRiT, including the image encoder, region proposal net, and text decoder head, is optimized during training, which costs a vast amount of training resources.
Our model only optimizes the lightweight feature mixer, which reduces the cost by lessening the memory consumption and bandwidth for syncing gradients.
2) The text decoding is initialized from scratch in GRiT, which constrains its language modeling ability due to the limited amount of region captioning data.
Whereas our method leverages pre-trained language models by mapping SAM's visual features into the language embedding space.
It raises two merits:
1) As the power of language model scales, we can observe improved performance on tests.
Our model with LLAMA-3B yields superior performance on the VG test set.
2) Since we do not finetune the language model to adapt new data distributions, it is possible to further improve our model based on the language aspect, \eg, chat-style interaction~\cite{pengInstructionTuningGPT42023_vicuna,liuVisualInstructionTuning2023c_llava}.
\cref{sec4:fig:qualitative} visualizes the predictions.

\begin{table}[t]
\centering
\caption{
The ablation of pretraining with weak supervision.
$\ast$: The model is trained solely on VG~\cite{krishnaVisualGenomeConnecting2016a_VG} for \textit{100K} steps.
$\dagger$: The model is first pre-trained for 100K, and then it is fine-tuned for 100K.
The training setting for ablations is different from that of~\cref{sec4:tab:comp}.
}
\label{sec4:tab:ablat-weak_sup}
\scalebox{0.97}{
\begin{tabular}{lccc} 
\toprule
Pretrain            & {C}                   & {M}                  & {S}                   \\ 
\hline
\textit{No Pretrain}$^\ast$         & 127.9                                 & 15.8                                 & 27.7                                  \\
\hline
COCO~\cite{linMicrosoftCOCOCommon2015a_mscoco} (img. 117K, cls. 80)$^\dagger$       & {\cellcolor[rgb]{0.996,1,0.698}}130.2 & {\cellcolor[rgb]{1,0.851,0.698}}16.0 & {\cellcolor[rgb]{0.996,1,0.698}}28.0  \\
V3Det~\cite{wangV3DetVastVocabulary2023a} (img. 183K, cls. 13K)$^\dagger$      & {\cellcolor[rgb]{1,0.851,0.698}}130.4 & {\cellcolor[rgb]{0.996,1,0.698}}16.0 & {\cellcolor[rgb]{1,0.851,0.698}}28.0  \\
O365~\cite{shaoObjects365LargeScaleHighQuality2019} (img. 1M, cls. 365)$^\dagger$ & {\cellcolor[rgb]{1,0.698,0.698}}134.5 & {\cellcolor[rgb]{1,0.698,0.698}}16.3 & {\cellcolor[rgb]{1,0.698,0.698}}28.7  \\
\bottomrule
\end{tabular}
}
\end{table}

\subsection{Ablation Study}

\begin{table}[t]
\centering
\caption{
    The ablation of training settings of the feature mixer and the text decoder.
    ``M.'': Feature mixer, ``T.D.'': Text decoder.
}
\label{sec4:tab:ablat-lr-text_dec}
\begin{tabular}{lllccc} 
\toprule
M. LR              & T.D.                   & {T.D. LR} & {C}                   & {M}                  & {S}                   \\ 
\hline
\multirow{5}{*}{1e-4} & \multirow{5}{*}{\begin{tabular}[c]{@{}l@{}} GPT2 \\-large \end{tabular}} & 5e-6                             & {\cellcolor[rgb]{0.996,1,0.698}}135.6 & {\cellcolor[rgb]{0.996,1,0.698}}16.3 & {\cellcolor[rgb]{0.996,1,0.698}}28.5  \\
                      &                             & 1e-6                             & 134.8                                 & 16.2                                 & 28.5                                  \\
                      &                             & 5e-7                             & 134.5                                 & 16.2                                 & 28.5                                  \\
                      &                             & 1e-7                             & {\cellcolor[rgb]{1,0.851,0.698}}135.6 & {\cellcolor[rgb]{1,0.851,0.698}}16.4 & {\cellcolor[rgb]{1,0.851,0.698}}28.8  \\
                      &                             & 0.0                             & {\cellcolor[rgb]{1,0.698,0.698}}136.0 & {\cellcolor[rgb]{1,0.698,0.698}}16.5 & {\cellcolor[rgb]{1,0.698,0.698}}28.9  \\
\hline
\multirow{5}{*}{5e-5}        & \multirow{5}{*}{\begin{tabular}[c]{@{}l@{}} GPT2 \\-large \end{tabular}}   & 5e-6                             & 129.1                                 & 15.7                                 & 27.5                                  \\
                             &                               & 1e-6                             & {\cellcolor[rgb]{0.996,1,0.698}}131.4 & 15.9                                 & 28.0                                  \\
                             &                               & 5e-7                             & 131.2                                 & {\cellcolor[rgb]{0.996,1,0.698}}16.0 & {\cellcolor[rgb]{0.996,1,0.698}}28.0  \\
                             &                               & 1e-7                             & {\cellcolor[rgb]{1,0.698,0.698}}132.5 & {\cellcolor[rgb]{1,0.698,0.698}}16.1 & {\cellcolor[rgb]{1,0.851,0.698}}28.2  \\
                             &                               & 0.0                             & {\cellcolor[rgb]{1,0.851,0.698}}131.7 & {\cellcolor[rgb]{1,0.851,0.698}}16.1 & {\cellcolor[rgb]{1,0.698,0.698}}28.2  \\
\hline
\multirow{5}{*}{1e-4}        & \multirow{5}{*}{GPT2}         & 5e-6                             & {\cellcolor[rgb]{0.996,1,0.698}}134.1 & {\cellcolor[rgb]{0.996,1,0.698}}16.2 & 28.4                                  \\
                             &                               & 1e-6                             & {\cellcolor[rgb]{1,0.698,0.698}}134.7 & {\cellcolor[rgb]{1,0.698,0.698}}16.3 & {\cellcolor[rgb]{0.996,1,0.698}}28.7  \\
                             &                               & 5e-7                             & {\cellcolor[rgb]{1,0.851,0.698}}134.5 & {\cellcolor[rgb]{1,0.851,0.698}}16.2 & {\cellcolor[rgb]{1,0.851,0.698}}28.7  \\
                             &                               & 1e-7                             & 133.2                                 & 16.1                                 & 28.6                                  \\
                             &                               & 0.0                             & 132.3                                 & 15.9                                 & {\cellcolor[rgb]{1,0.698,0.698}}28.9  \\
\hline
\multirow{5}{*}{5e-5}        & \multirow{5}{*}{GPT2}         & 5e-6                             & {\cellcolor[rgb]{1,0.698,0.698}}131.3 & {\cellcolor[rgb]{1,0.851,0.698}}16.0 & 28.0                                  \\
                             &                               & 1e-6                             & {\cellcolor[rgb]{1,0.851,0.698}}131.1 & {\cellcolor[rgb]{1,0.698,0.698}}16.0 & {\cellcolor[rgb]{1,0.851,0.698}}28.1  \\
                             &                               & 5e-7                             & {\cellcolor[rgb]{0.996,1,0.698}}130.6 & {\cellcolor[rgb]{0.996,1,0.698}}15.9 & {\cellcolor[rgb]{0.996,1,0.698}}28.1  \\
                             &                               & 1e-7                             & 130.4                                 & 15.9                                 & {\cellcolor[rgb]{1,0.698,0.698}}28.2  \\
                             &                               & 0.0                             & 126.3                                 & 15.4                                 & 27.9                                  \\
\bottomrule
\end{tabular}
\end{table}

In the early stage of experimenting, we spent less computational budgets to validate the efficiency of different design choices.
Specifically, for all models in this section, we constrained the budges to 8 16GB V100 GPUs with a batch size of 8.
By default, the models are trained solely on VG~\cite{krishnaVisualGenomeConnecting2016a_VG} \textit{without} data augmentation for 200K steps.

\noindent \textbf{The effectiveness of weak supervision pre-training}.
To preliminarily validate the effectiveness of pretraining with weak supervision.
We leveraged three object detection datasets:
1) MS COCO~\cite{linMicrosoftCOCOCommon2015a_mscoco} contains about 117K images and 80 classes;
2) V3Det~\cite{wangV3DetVastVocabulary2023a} is a rich-semantic detection dataset with around 183K images and 13k classes;
3) Objects365~\cite{shaoObjects365LargeScaleHighQuality2019} is a large-scale detection dataset with over 1M images, 27M regions, and 365 class labels.
The model was first pre-trained for 100K steps and finetuned for another 100K without other modifications.
We set another baseline trained directly on VG for 100K steps.
\cref{sec4:tab:ablat-weak_sup} presents that the pretraining with concept labels can facilitate the convergence of training on VG,
and the larger the scale of the images, the better the test performance.
Under a similar amount of samples, an increase in class labels can slightly improve performance.
The finding encourages us to further enlarge the pretraining data scale in the future.

\noindent \textbf{The hyper-parameters of the text decoder and the feature mixer}.
To determine the training recipe for the text decoder, we experimented with two factors: 1) The size of the text decoder; and 2) The optimization of the text decoder.
We tried two variants of GPT2 transformer decoder models~\cite{radfordLanguageModelsAre_gpt2}, which are GPT2-large with 774M parameters and GPT2 with 127M.
They are all from the official release which are trained on WebText dataset~\cite{radfordLanguageModelsAre_gpt2}.
Given the different learning rates of the feature mixer, we then tested different learning rates (\ie, 0.0, 1e-7, 5e-7, 1e-6, 5e-6) for the text decoder.

Two conclusions can be drawn from \cref{sec4:tab:ablat-lr-text_dec}.
1) The feature mixer requires a relatively large learning rate to converge to good performance.
2) When the text decoder is small (GPT2 with 127M), we need to finetune it to achieve better results.
In contrast, using a larger text decoder like GPT2-large (774M), finetuning may impede the performance, and fixing the decoder can yield even better scores compared with the small one.

We chose a large text decoder without any finetuning in this paper given the considerations of both capacity and efficiency.
In this, we not only keep the knowledge inside the language model for future improvement,
but enable low-cost training of the model.

\noindent \textbf{The size of feature mixer}.
The additional feature mixer for text decoding is a bi-directional transformer,
which fuses the query and prompt tokens with self-attention, and the image tokens with cross-attention.
The original one used in~\cite{kirillovSegmentAnything2023b_SAM} is a two-layer one that is highly computational-efficient with solely 3M parameters.

To investigate how the size of the feature mixer affects the extracted region feature,
we test a different number of layers for the additional transformer, ranging from 2 with 3M parameters to 24 with 30M parameters.

\cref{sec4:tab:mixer_size} demonstrates the final scores on the VG test split.
As the number of layers increases, the n-gram metrics ramp up as well.
Only until 12 layers do its performance reach the peak,
then adding more layers harms the performance.
Noticeably, \cite{liBLIP2BootstrappingLanguageImage2023_blip2} used 12 layers feature mixer to extract prominent features for image captioning,
which has over 105M parameters.
While ours only consists of 19.4M.

\begin{table}[t]
\centering
\caption{
The effect of different number of layers in the feature mixer.
Note that this is the \textit{only} trainable module in our models.
}
\label{sec4:tab:mixer_size}
\begin{tabular}{llccc}
\toprule
{\# of Layers} & \# of Params & {C}                   & {M}                  & {S}                   \\ 
\hline
2 & \ \ 3.3 M                                  & 108.8                                 & 13.6                                 & 24.6                                  \\
4 & \ \ 6.5 M                                  & 109.8                                 & 14.0                                 & 25.6                                  \\
8 & 12.8 M                                  & {\cellcolor[rgb]{1,0.851,0.698}}127.0 & {\cellcolor[rgb]{1,0.851,0.698}}15.3 & {\cellcolor[rgb]{1,0.851,0.698}}27.8  \\
12 & 19.1 M                                 & {\cellcolor[rgb]{1,0.698,0.698}}127.7 & {\cellcolor[rgb]{1,0.698,0.698}}15.3 & {\cellcolor[rgb]{1,0.698,0.698}}27.9  \\
24 & 38.0 M                                 & {\cellcolor[rgb]{0.996,1,0.698}}124.5 & {\cellcolor[rgb]{0.996,1,0.698}}15.0 & {\cellcolor[rgb]{0.996,1,0.698}}27.3  \\
\bottomrule
\end{tabular}
\end{table}

\noindent \textbf{The architecture of feature mixer}.
We experiment with four major architectures, which are 1) the one with ROI-Align operator~\cite{heMaskRCNN2018a}, which is described in the main paper; 2) the one that directly decoding the query tokens from SAM; 3) the one that does not rely on the fused tokens from SAM's feature mixer (in other words, not reusing SAM's tokens); 4) the one that utilizes the query tokens from SAM to decode texts.
To make the ROI align one stronger, we add an MLP stated in~\cite{liu2023improved_llava_1_5}, which is a two-layer MLP with GELU activation~\cite{hendrycks2016gaussian_gelu}.

\cref{sec4:tab:model_arch} shows that query-based mixers perform significantly better than those using ROI-align,
indicating the effectiveness of progressive feature aggregation.
Directly decoding SAM's query token restricts the capacity of the mixer.
Incorporating additional query tokens for captioning boosts the performance of the model.
Moreover, resuing the features of SAM further improves the captioning results.

\begin{table}[t]
\centering
\caption{The ablation of feature mixer design.}
\label{sec4:tab:model_arch}
\begin{tabular}{lccc} 
\toprule
Method                     & {C}                   & {M}                  & {S}                   \\ 
\hline
ROI Align~\cite{heMaskRCNN2018a}                   & 45.2                                  & 9.4                                  & 11.6                                  \\
ROI Align + MLP~\cite{liu2023improved_llava_1_5}             & 82.5                                  & 12.1                                 & 19.3                                  \\
\hline
SAM Query~\cite{kirillovSegmentAnything2023b_SAM}                  & {\cellcolor[rgb]{0.996,1,0.698}}130.6 & {\cellcolor[rgb]{0.996,1,0.698}}15.9 & {\cellcolor[rgb]{0.996,1,0.698}}28.4  \\
Text Query~ w/o SAM Tokens & {\cellcolor[rgb]{1,0.851,0.698}}136.6 & {\cellcolor[rgb]{1,0.851,0.698}}16.4 & {\cellcolor[rgb]{1,0.851,0.698}}29.2  \\
Text Query w/ SAM Tokens   & {\cellcolor[rgb]{1,0.698,0.698}}137.4 & {\cellcolor[rgb]{1,0.698,0.698}}16.5 & {\cellcolor[rgb]{1,0.698,0.698}}29.3  \\
\bottomrule
\end{tabular}
\end{table}

\noindent \textbf{The size of the SAM image encoder}.
We investigate how different SAM encoders may affect the captioning performance,
by testing the three official encoders from~\cite{kirillovSegmentAnything2023b_SAM},
which are three ViT~\cite{dosovitskiyImageWorth16x162021a_vit} with different scale:
base, large, and huge.
Surprisingly, different size of the SAM image encoders results in similar final performance.
We chose the ViT huge as the default image encoder as it performs slightly better than others.

\begin{table}[t]
\centering
\caption{The ablation of using different sizes of image encoder.}
\label{sec4:tab:sam_size}
\begin{tabular}{llccc} 
\toprule
Method & \# of Params         & {C}                   & {M}                  & {S}                   \\ 
\hline
SAM-ViT-base & \ \ 86M  & {\cellcolor[rgb]{1,0.851,0.698}}130.2 & {\cellcolor[rgb]{1,0.698,0.698}}16.0 & {\cellcolor[rgb]{0.996,1,0.698}}28.2  \\
SAM-ViT-large & 307M & {\cellcolor[rgb]{0.996,1,0.698}}129.6 & {\cellcolor[rgb]{0.996,1,0.698}}15.9 & {\cellcolor[rgb]{1,0.851,0.698}}28.3  \\
SAM-ViT-huge & 632M  & {\cellcolor[rgb]{1,0.698,0.698}}130.9 & {\cellcolor[rgb]{1,0.851,0.698}}16.0 & {\cellcolor[rgb]{1,0.698,0.698}}28.5  \\
\bottomrule
\end{tabular}
\end{table}

\noindent \textbf{The efficacy of data augmentation}.
We found that with an enlarged batch size in multiple-node training, the model experienced an overfitting problem which led to inferior test performance.
To fight against the problem, we resort to strong augmentation from~\cite{ghiasiSimpleCopyPasteStrong2021a_copy_paste}, the large-scale jittering.
\cref{sec4:tab:data_aug} demonstrates that using the strong augmentation not only alleviates the overfitting problem but enhances the model's performance.

\section{Conclusions and Discussions}

We preliminarily demonstrate a regional captioning system by adapting a powerful class-agnostic segmentation model, SAM~\cite{kirillovSegmentAnything2023b_SAM},
with a  lightweight (typically in the order of tens of millions) query-based feature mixer that bridges SAM with the language model.
The mixer is the only optimizable module thus the training is both \textit{faster} and \textit{scalable},
as it costs less computation, less memory usage, and less communication bandwidth.
To better generalize our model, we pre-train the system with weak supervision which transfers the general knowledge of the visual concepts beyond the limited regional captioning data, Visual Genome (VG)~\cite{krishnaVisualGenomeConnecting2016a_VG}.
We extensively validate our design choices and evaluate our method, demonstrating its strong performance.

\noindent \textbf{Limitations}.
1) Wrong attribute prediction. \eg, the models could predict the wrong colors or textures;
2) Distinguishing similar visual concepts. \eg, the model may confuse ``lemon'' with ``orange'';
3) Alignment with mask predictions: As we do not supervise the alignment, the model may predict mask and captions for the fore- and background separately.
The drawbacks, \textit{esp.} 1) and 2), may be addressed by weak supervision and self-training~\cite{betkerImprovingImageGeneration_dalle3}.

\begin{table}[t]
\centering
\caption{
    The ablation of using data augmentation.
    ``LM'': Language model, ``Aug.'': Augmentation.
}
\label{sec4:tab:data_aug}
\begin{tabular}{llccc}
\toprule
LM                          & Aug.          & {C}                   & {M}                  & {S}                   \\ 
\hline
\multirow{3}{*}{GPT2-large} & No LSJ       & {\cellcolor[rgb]{0.996,1,0.698}}137.6 & {\cellcolor[rgb]{0.996,1,0.698}}16.5 & {\cellcolor[rgb]{0.996,1,0.698}}29.3  \\
                            & LSJ (1.0, 2.0) & {\cellcolor[rgb]{1,0.851,0.698}}140.2 & {\cellcolor[rgb]{1,0.698,0.698}}16.7 & {\cellcolor[rgb]{1,0.851,0.698}}29.9  \\
                            & LSJ (0.1, 2.0) & {\cellcolor[rgb]{1,0.698,0.698}}140.8 & {\cellcolor[rgb]{1,0.851,0.698}}16.7 & {\cellcolor[rgb]{1,0.698,0.698}}29.9  \\
\hline
\multirow{3}{*}{LLAMA-3B} & No LSJ       & {\cellcolor[rgb]{0.996,1,0.698}}137.7 & {\cellcolor[rgb]{0.996,1,0.698}}16.4 & {\cellcolor[rgb]{0.996,1,0.698}}29.2  \\
                          & LSJ (1.0, 2.0) & {\cellcolor[rgb]{1,0.851,0.698}}142.1 & {\cellcolor[rgb]{1,0.851,0.698}}16.7 & {\cellcolor[rgb]{1,0.851,0.698}}30.0  \\
                          & LSJ (0.1, 2.0) & {\cellcolor[rgb]{1,0.698,0.698}}142.6 & {\cellcolor[rgb]{1,0.698,0.698}}16.8 & {\cellcolor[rgb]{1,0.698,0.698}}30.1  \\
\bottomrule
\end{tabular}
\end{table}

\noindent \textbf{Weak supervision and self-training}.
We only leverage 1.8M weak supervision data~\cite{shaoObjects365LargeScaleHighQuality2019,linMicrosoftCOCOCommon2015a_mscoco} to complement the regional captioning data, VG~\cite{krishnaVisualGenomeConnecting2016a_VG}.
Our ablation about the effectiveness of weak supervision
shows that the scale of images matters more than the variety of labels,
which is intuitive as we want the model to align and \textit{generalize} as much as visual concepts with the language models.
Thus pertaining the model with \textit{bigger datasets} like~\cite{kuznetsovaOpenImagesDataset2020_openimage} may lead to better generalizability.
Another approach to leverage \textit{image captioning} data as in~\cite{zou2023generalized_xdecoder,liGroundedLanguageImagePretraining2022a_glipv1},
but it requires to solve the problem of granularity mismatching~\cite{zou2023generalized_xdecoder}.
\textit{Self-training} is the ultimate goal that could scale both the data and the generalizability of the model.
It demonstrates effectiveness in image captioning~\cite{liBLIPBootstrappingLanguageImage2022_blip}, segmentation~\cite{kirillovSegmentAnything2023b_SAM}, open-vocabulary detection~\cite{mindererScalingOpenVocabularyObject2023a_owlvitv2}, and text-to-image generation~\cite{betkerImprovingImageGeneration_dalle3}.
We believe this work serves as a footstone towards scaling regional captioning data~\cite{kirillovSegmentAnything2023b_SAM,betkerImprovingImageGeneration_dalle3,mindererScalingOpenVocabularyObject2023a_owlvitv2} in the future.

\noindent \textbf{Insight of lifting SAM for regional captioning}.
Although there are no semantic labels in the training data, SAM still implies \textit{high-level semantics} that are sufficient for captioning.
The masks used to train SAM are labeled in a way where annotators are asked to draw masks for every \textit{things} or \textit{stuff} they recognized~\cite{kirillovSegmentAnything2023b_SAM}.
After several rounds of self-training and bootstrapping the data to 1B masks, the attained models possess implicit general knowledge about the visual world.
Therefore, we can \textit{align} the implicit general knowledge with natural languages to caption regions.
We believe this work sheds light on exploring the emerging ability~\cite{Wei2022EmergentAO_emerge_ability_lm} in vision from low-level data or pre-trains.

{
    \small
    \bibliographystyle{ieeenat_fullname}
    \bibliography{sca.updated.revise_key}
}


\end{document}